\documentclass[runningheads]{llncs}
\usepackage{bbding}
\usepackage{graphicx}
\usepackage{xcolor}
\usepackage{textcomp}
\usepackage{algorithm}
\usepackage{arevmath}     
\usepackage[noend]{algpseudocode}
\usepackage{graphicx}
\usepackage{diagbox}
\usepackage{slashbox}
\begin{document}
\title{An Efficient K-means Clustering Algorithm for Analysing COVID-19}



\authorrunning{Zubair et al.}

\author{Md. Zubair$^{1,a}$ \and
MD.Asif Iqbal$^{1,b}$ \and Avijeet Shil$^{1,c}$ \and Enamul Haque$^{2,d}$ \and Mohammed Moshiul Hoque$^{1,e}$ \and Iqbal H. Sarker$^{1,*}$}
%
%
\institute{$^1$ Dept of Computer Science \& Engineering, Chittagong University of Engineering \& Technology, Chittagong-4349, Bangladesh.\\
$^2$ McMaster University, Canada.\\
\email{$^{a,b,c}$\{zubairhossain773,asifiqbalsagor123,avijeetshil110\}@gmail.com, $^d$enamul.haque@uwaterloo.ca, $^e$moshiul\_240@cuet.ac.bd,\\
$*$Correspondence:iqbal@cuet.ac.bd}
}

%
\maketitle   
\vspace{-2mm}
\begin{abstract}

COVID-19 hits the world like a storm by arising pandemic situations for most of the countries around the world. The whole world is trying to overcome this pandemic situation. A better health care quality may help a country to tackle the pandemic. Making clusters of countries with similar types of health care quality provides an insight into the quality of health care in different countries. In the area of machine learning and data science, the K-means clustering algorithm is typically used to create clusters based on similarity. In this paper, we propose an \textit{efficient K-means clustering} method that determines the initial centroids of the clusters efficiently. Based on this proposed method, we have determined \textit{health care quality clusters} of countries utilizing the COVID-19 datasets. Experimental results show that our proposed method reduces the number of iterations and execution time to analyze COVID-19 while comparing with the traditional k-means clustering algorithm.    

\vspace{-2mm}
\end{abstract}

\section{Introduction}
\vspace{-2mm}
COVID-19 was detected near the end of 2019 from the Chinese city of Wuhan \cite{world2020coronavirus}. The virus spread rapidly by transmission throughout the whole world within a short period of time. The World Health Organization (WHO) had to declare COVID-19 pandemic. To eradicate the pandemic situation, different countries have taken initiatives to remold their health care quality. Clustering the countries with similar types of health care quality helps us to know about the health care quality comparing with the rest of the counties.

In the area of machine learning and data science, both supervised learning and unsupervised learning are popular to solve various kinds of real world problems \cite{sarker2020mobile}  \cite{sarker2020cybersecurity}. In the case of clustering, unsupervised machine learning algorithms are being used \cite{sarker2019context} \cite{han2011data}. The unsupervised machine learning algorithm typically identifies insight structures of the data from unlabelled data contained in the dataset. The clustering algorithm finds and divides the data points according to the similarity of the hidden structures of the dataset \cite{sarker2018individualized}. K-Means \cite{han2011data} clustering is one of the most important and widely used unsupervised machine learning algorithms as it is widely used to identify the hidden structure automatically.

The algorithm is used to find the number of clusters so that it is possible to divide the unlabelled dataset into subgroups. It is done by calculating the distances from a centroid of a cluster. K-means algorithm is an NP-Hard problem \cite{vattani2009hardness}. Efficiently estimating the initial centroids is a difficult problem. At the initial state, we need to fix the coordinates of the initial centroid for finding the number of clusters. In the traditional K-Means clustering algorithm, initial centroids usually being selected randomly. Thus, determining the initial coordinate points of centroids can play a signifcant role in clustering, in which we are interested in. The key contributions of our work are as follows -
\begin{description}
  \item[$\bullet$] Our proposed method efficiently selects the k-means clustering algorithm’s centroids that provides an optimum number of constant iteration as well as execution time. 
  \item[$\bullet$] We have clustered similar types of countries according to the health care quality during the COVID-19 pandemic. 
  \item[$\bullet$] We have tested our model with a COVID-19 real-world dataset to show the efficiency of our model with the existing models.
\end{description}
  
In the following, several related works have been discussed in Section \ref{related}. We present our proposed methodology in Section \ref{method}. In Section \ref{experiment}, we have shown the experimental results and comparative analysis. In Section \ref{discussion} we highlight some key points and concludes this paper.

\vspace{-2mm}
\section{Related Work}
\label{related}
\vspace{-1mm}
Several approaches have been proposed to find the initial cluster centroids more efficiently. In this section, we bring some of these works. M. S. Rahman et al. \cite{rahim2017initial}, provides a centroids selection method based on radial and angular coordinates. However, the number of iterations of his proposed idea is not constant for all the instances. Also, the execution time of that method increases drastically by the increase of the cluster number.
A. Kumar et al. also proposed to find initial centroids based on the dissimilarity tree \cite{kumar2015new}. Although this method improves k-means clustering, the execution time is not exalted significantly. Also, the smaller datasets that are used for experiment results can not provide insight into a large dataset.
In \cite{mahmud2012improvement}, M.S. Mahmud et al. proposed a novel approach to find the initial centroids by calculating the mean of each distance of the data points. It only describes 3 clusters of given three datasets with the execution time. Improvement of execution time is also trivial. The concept is based on the weighted average.
In \cite{goyal2014improving}, another approach is made to find the initial centroids efficiently. Here in M. Goyal et al. also tried to find the centroids by dividing the sorted distances with k, the number of equal partitions. No execution time was given for this proposed method. 
S. Na et al. \cite{na2010research} have proposed the use of two elementary data structures to store cluster labels and the distance of all data items with each iteration. On the next iteration, data of the previous iteration was used. But, execution time was trivial for the dataset experimented by authors.
M. A. Lakshmi et al. \cite{lakshmi2019initial} have proposed a method to find initial centroids by using the nearest neighbor method. They compared their idea by using SSE(Sum of the Squared Differences) with random and kmeans++ initial selection. Their SSE is roughly similar to random and kmeans++ initial selection. Moreover, They did not provide any comparison concerning execution time as well. 


S. R. Vadyala et al. proposed a combined algorithm with k-means and LSTM to predict the number of confirmed cases of COVID-19 \cite{vadyala2020prediction}. LSTM is abbreviated as long short-term memory, an artificial recurrent neural network architecture used for Deep learning.
In \cite{poompaavai2019clustering}, author A. Poompaavai et al. attempted to identify the affected areas by COVID-19 of India by using the k-means clustering algorithm.
Many approaches related to COVID-19 problem, k-means clustering has been used. In \cite{sonbhadra2020target},  S.K.  Sonbhadra et al. proposed a novel bottom-up approach for COVID-19 articles using k-means clustering along with DBSCAN and HAC.

\vspace{-2mm}
\section{Proposed Methodology}
\vspace{-2mm}
\label{method}
In this section, we present our k-means clustering-based COVID-19 analysis to determine the clusters according to the health care quality of the countries. The result of the k-means clustering is effected on the selection or assignment of initial centroids \cite{berkhin2006survey}. So, it is necessary to select the centroids more systematically to improve the performance of the k-means clustering algorithm, also the execution time. In our approach, we use Principal Component Analysis (PCA) \cite{sarker2020contextpca} while determining the centroids.

\begin{figure}[!hbt]
\vspace{-2mm}
\centering
    \includegraphics[width = \linewidth]{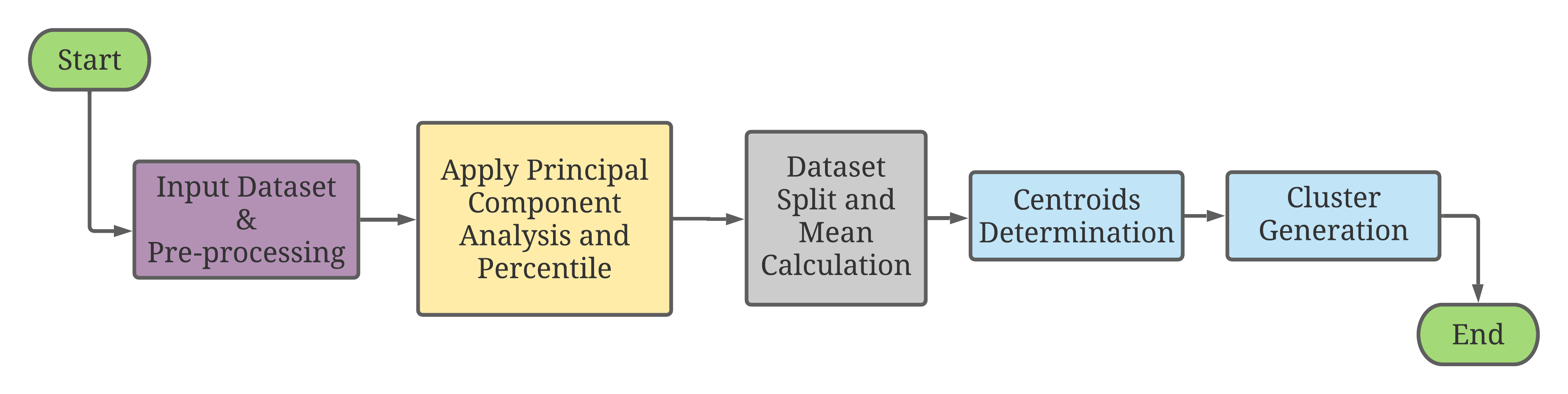}\par 
\caption{Flowchart of the proposed method}
\label{fig:methodology}
\vspace{-6mm}
\end{figure}

The following flowchart in Figure \ref{fig:methodology} depicts the overall process of our proposed method. Our proposed method can determine the coordinates of initial centroids which can obtain the coordinates more precisely than the existing methods.

\vspace{-6mm}
\subsection{Input Dataset}
\label{input}
Many machine learning algorithms, including both supervised and unsupervised methods have been applied to the COVID-19 dataset \cite{poompaavai2019clustering}. Each attribute of the dataset must contain numerical values for approaching the k-means algorithm \cite{han2011data}. If not then, some pre-processing might be required for handling missing values and non-numeric values. Before going to further methodology, it needs to be ensured that each attributes values in numeric.

For creating our model, we have used a few datasets for selecting the features, required for analyzing the health care quality of the countries. The selected datasets are owid-covid-data \cite{TotalCOV22:online}, covid-19-testing-policy \cite{COVID19T92:online}, public-events-covid \cite{COVID19T92:online}, covid-containment-and-health-index \cite{COVID19T92:online}, inform-covid-indicators\cite{UNCOVERC93:online}. It is worth mentioning that we used the data up to 11\textsuperscript{st} August 2020. For instance, some of the attributes of the owid-covid-data \cite{TotalCOV22:online} are shown in Table \ref{tab:1}. 
\begin{table}[]
\vspace{-4mm}
\caption{Sample data of COVID-19 dataset }
\label{tab:1}
\begin{tabular}{|c|c|c|c|c|c|c|c|c|c|c}
\hline
Country  & Date & \begin{tabular}[c]{@{}c@{}}total\\ cases\\ per\\ million\end{tabular} & \begin{tabular}[c]{@{}c@{}}new\\ cases\\ per \\ million\end{tabular} & \begin{tabular}[c]{@{}c@{}}total\\ deaths \\ per \\ million\end{tabular} & \begin{tabular}[c]{@{}c@{}}new\\ deaths\\ per\\ million\end{tabular} & \begin{tabular}[c]{@{}c@{}}cardiovasc\\ death\\ rate\end{tabular} & \begin{tabular}[c]{@{}c@{}}hospital\\ beds\\ per\\ thousand\end{tabular} & \begin{tabular}[c]{@{}c@{}}life\\ expectancy\end{tabular} \\ \hline
Australia & 11/08/20 & 839.102 & 12.275 & 12.275& 0.706 & 107.791 & 3.84 & 83.44 \\ \hline
Bangladesh & 11/08/20 & 1581.808 & 17.651 & 20.876 & 0.237 & 298.003 & 0.8 & 72.59\\\hline
China & 11/08/20 & 61.769 & 0.079 & 3.258 & 0.003 & 261.899 & 4.34 & 76.91 
\\\hline

\end{tabular}
\vspace{-4mm}
\end{table}
Covid-19-testing-policy \cite{COVID19T92:online} dataset contains the categorical values of the testing policy of the countries shown in table \ref{tab:2}.

\begin{table}[]
\vspace{-4mm}
\caption{Sample data of  Covid-19-testing-policy }
\label{tab:2}
\centering
\begin{tabular}{|c|c|c|c|}
\hline
Entity & Code & Date & Testing Policy \\ \hline
Australia & AUS & Aug 11, 2020 & 3 \\ \hline
Bangladesh & BGD &Aug 11, 2020 & 2 \\ \hline
China & CHN &Aug 11, 2020 & 3 \\ \hline

\end{tabular}
\vspace{-4mm}
\end{table}

Other datasets also contains such types of features which is required for ensuring the health care quality of a country. These real-world datasets help us to analyze our proposed method for real-world scenarios.

\subsection{Principal Component Analysis and Percentile}

Principal component analysis (PCA) is a method, that utilizes an orthogonal transformation to translate a set of observations of potentially correlated variables into a set of values of linearly uncorrelated variables that are called principal components. PCA is widely used in data analysis and for making a predictive model considering dimensionality reduction \cite{abdi2010principal} \cite{sarker2020contextpca}, that has been taken into account in our approach.



On the other hand, the percentile method \cite{altman1994statistics} used in our model divides the whole dataset into 100 different parts. Each part contains 1 percent data of the total dataset. For example, the 25th percentile means this part contains 25 percent data of the total dataset. That implies, using the percentile method, we can split our dataset into different distributions according to our given values. The percentile formula is given below \cite{altman1994statistics}:
				\[R = \frac{\rho}{100} *(\eta+1)\]
Here, $\rho$ = The Percentile wants to find, $\eta$ = Total Number of values, R = Percentile at $\rho$.

\subsection{Dataset Split and Mean Calculation}
\label{split and mean}

After splitting the reduced dimensional dataset through percentile, we then extract the split data from the primary dataset by indexing for each percentile. In this process, we get back the original data. After retrieving the original data for each percentile, we have calculated the mean of each attribute. These means from the split dataset is the initial centroids of our proposed method.

\subsection{Centroids Determination}
After splitting the dataset and calculating the mean according to the subsection 3.3, we have selected the means of each split dataset as a centroid. These centroids are taken into account as initial centroids for the efficient k-means clustering algorithm. 

\subsection{Cluster Generation}

At the last step, we have executed our modified k-means algorithm until the centroids converge. Passing our proposed centroids instead of random or kmeans++ centroids through the k-means algorithm we have generated the final clusters \cite{arthur2006k}. The proposed method always considers the same centroids for each test.
The pseudocode of our whole proposed methodology is given in the algorithm \ref{algo3}. In the next section, evaluation and experimental results are discussed.

\algdef{SE}[PROCEDURE]{Procedure}{EndProcedure}%
   [2]{\algorithmicprocedure\ \textproc{#1}\ifthenelse{\equal{#2}{}}{}{(#2)}}%
   {\algorithmicend\ \algorithmicprocedure}%
\algdef{SE}[FUNCTION]{Function}{EndFunction}%
   [2]{\algorithmicfunction\ \textproc{#1}\ifthenelse{\equal{#2}{}}{}{(#2)}}%
   {\algorithmicend\ \algorithmicfunction}%

\begin{algorithm}
\caption{Proposed Methodology}
\label{algo3}
\begin{algorithmic}[1]
\item Input: A dataset \textbf{D} and Number of clusters \textbf{K}
\item Output: Efficient initial centroids for \textbf{K} clusters
 \Procedure{Proposed Method}{D}       
    \State All n attributes ${a1,a2,a3,...,an}$ of \textbf{D} must be numeric. If there is any non-numeric attribute just convert it to numeric value.
    \State Apply Principal Component Analysis (PCA) with 2 components to the dataset, \textbf{D}.
    \State Apply percentile for splitting the whole dataset into \textbf{K} equal parts based on 1st component.
    \State Extract the split dataset from primary data by index.
    \State Calculate the mean of each attribute of the split datasets.
    \State Take the mean of each dataset as the initial clusters centroids, $C={c1,c2,...,ck}$, where $c1,c2,..,ck$ are the initial centroids for $1st,2nd,....,k$ clusters consecutively.
    \State Assign the centroids to the k-means clustering algorithm
    \State Applying k-means algorithm with proposed initial centroids.
\EndProcedure
\end{algorithmic}
\end{algorithm}

\section{Implementation and Experimental Results Analysis}
\label{experiment}
To measure the effectiveness and validate our proposed model for selecting the optimum initial centroids for the k-means clustering algorithm, we have implemented and showed experimental results with real-world dataset. We have used five COVID-19 datasets and merged them to have a handful of features for clustering the countries according to their health quality during COVID-19.



\subsection{Datasets Pre-processing}
\vspace{-2mm}

We have merged the datasets described in subsection \ref{input} according to the country name. With the regular expression, some pre-processing and data cleaning have been conducted in the case of merging the data. We have also handled some missing data consciously. There are so many attributes regarding COVID-19, among them 25 attributes had been selected finally, as these attributes closely signify the health care quality of a country. The attributes represent categorical and numerical values. These are country name, cancellation of public events (due to public health awareness), stringency index\footnote{It is one of the matrices used by Oxford COVID-19 Government Response Tracker \cite{Coronavi44:online}. It delivers a picture of the country's enforced strongest measures.}, testing policy ( category of testing facility available to the mass people), total positive case per million, new cases per million, total death per million, new deaths per million, cardiovascular death rate, hospital beds available per thousand, life expectancy, inform the COVID-19 risk (rate), hazard and exposure dimension rate, people using at least basic sanitation services (rate), inform vulnerability(rate), inform health conditions (rate), inform epidemic vulnerability (rate), mortality rate, prevalence of undernourishment, lack of coping capacity, access to healthcare, physicians density, current health expenditure per capita, maternal mortality ratio. All of the attributes are closely investigated before feeding the model.

As we are going to make clusters for the countries with similar types of healthcare quality, the optimum number of clusters is 4, defined by the elbow method \cite{kodinariya2013review}.

\vspace{-2mm}
\subsection{Centroids of COVID-19 Dataset}
\begin{figure}[!hbt]
\centering
    \includegraphics[width = 0.6 \linewidth, height = 5cm]{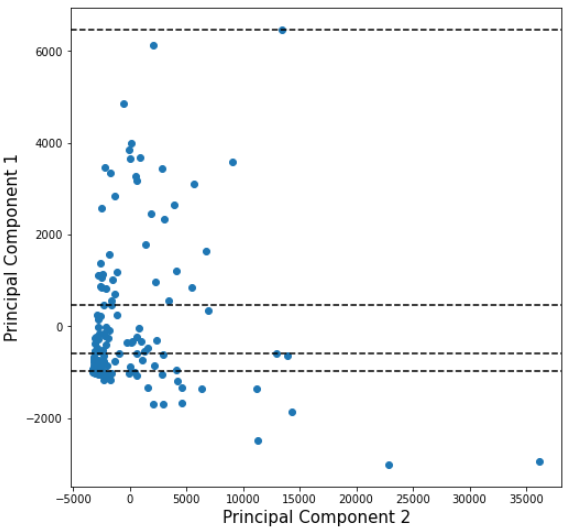}\par 
\caption{PCA with Percentile on COVID-19 Dataset}
\label{fig:pcacar}
\vspace{-3mm}
\end{figure}

\vspace{-2mm}
We have applied Principal Component Analysis (PCA) to convert a high dimensional dataset into two dimensions. The number of clusters is 4 for the COVID-19 dataset. After applying PCA, we have used percentile concepts to divide the whole dataset into K equal parts. For K=4, each portion contains 25\% of the dataset. For this purpose, we have calculated the percentile for 25\%, 50\%, 75\%, and 99.9\% data. We have divided the data horizontally because it provides a good intuition to the cluster. Figure \ref{fig:pcacar} depicts plotting the data with two dimensional PCA where horizontal lines represents the splitting according to percentile.

As described in subsection \ref{split and mean}, after splitting the dataset and mean calculation we have got the proposed initial centroids of k-means. Efficient centroids of COVID-19 for 4 centroids are given in table \ref{tab:3}. Here only 7 attributes from 25 attributes are shown for demonstration purposes. In table \ref{tab:3}, \emph{c1,c2,c3 and c4} represent initial clusters centroids consecutively.
\vspace{2mm}

\begin{table}[]
\vspace{-6mm}
\caption{Initial Centroids of COVID-19 Dataset}
\label{tab:3}
\begin{tabular}{|c|c|c|c|c|c|c|c|}
\hline
\backslashbox{Clusters\\Centroids}{Columns }& \begin{tabular}[c]{@{}c@{}}total\\ cases\\ per\\ million\end{tabular} & \begin{tabular}[c]{@{}c@{}}new\\ cases\\ per \\ million\end{tabular} & \begin{tabular}[c]{@{}c@{}}total\\ deaths \\ per \\ million\end{tabular} & \begin{tabular}[c]{@{}c@{}}new\\ deaths\\ per\\ million\end{tabular} & \begin{tabular}[c]{@{}c@{}}cardiovasc\\ death\\ rate\end{tabular} & \begin{tabular}[c]{@{}c@{}}hospital\\ beds\\ per\\ thousand\end{tabular} & \begin{tabular}[c]{@{}c@{}}life\\ expectancy\end{tabular} \\ \hline
c1 & 5052.931 & 43.53 & 94.81 & 1.06 & 277.75 & 1.6 & 67.7 \\ \hline
c2 & 1456.5 & 17.02 & 28.69 & 0.23 & 320.57 & 2.361 & 67.18 \\ \hline
c3 & 2190.72 & 33.01 & 47.76 & .81 & 270.99 & 3.16 & 74.92 \\ \hline
c4 & 3577.2 & 27.41 & 17.57 & .28 & 163.93 & 4.36 & 80.73 \\ \hline
\end{tabular}
\vspace{-2mm}
\end{table}
\vspace{-6mm}

\subsection{Evaluation Process}
In machine learning and data science, computational power is one of the main issues. Because at a time, a computer needs to process a large amount of data. So, reducing computational cost is a big deal.
As discussed in the methodology in section \ref{method}, we implemented our proposed method. Though many researchers proposed many ideas those are discussed in the related work section. We have compared our method with the best existing kmeans++  method \cite{arthur2006k}. We have measured the effectiveness of the model with 
\begin{itemize}
    \item[$\bullet$] Number of iterations needed for finding the final clusters 
    \item[$\bullet$] Execution time for reaching out to the final clusters
\end{itemize}

These two things are compared in the upcoming subsections.

\subsection{Efficiency analysis }
\begin{figure}[!t]
\centering{\includegraphics[width=\linewidth]{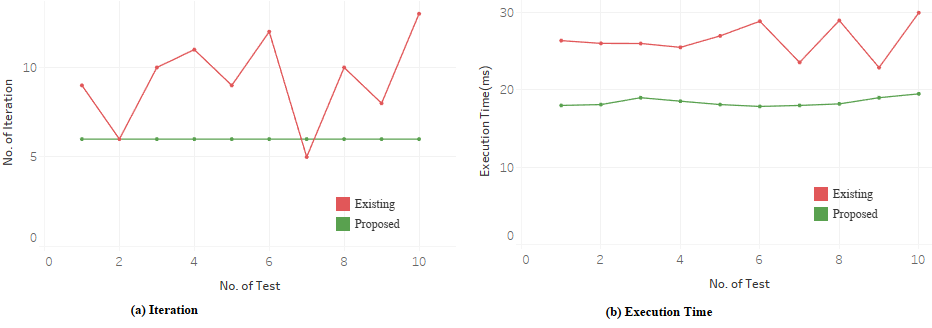}} 
\caption{Experimental result for Iteration and Execution time with COVID-19 dataset}
\label{fig:3}
\vspace{-2mm}
\end{figure}

\begin{figure}[!hbt]
\vspace{-4mm}
\centering
    \includegraphics[width = \linewidth, height = 5cm]{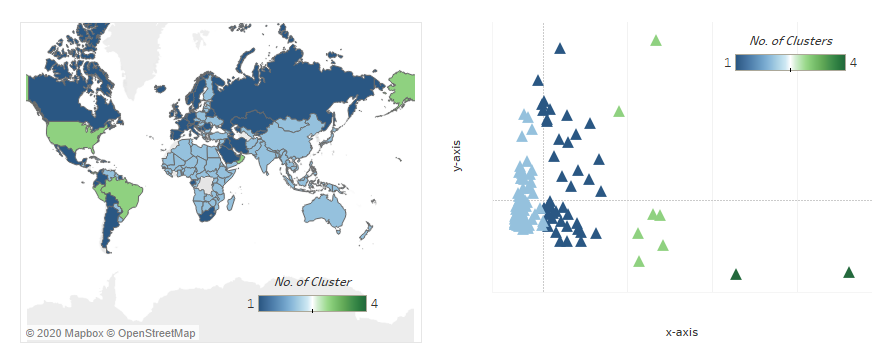}\par 
\caption{Health care quality clusters with a) real-world map and b) 2D cluster plot}
\label{fig:4}
\vspace{-4mm}
\end{figure}

In figure \ref{fig:3}, we show the experimental result of the COVID-19 dataset for 10 tests. In the graph, the green and the red line represents results for the proposed and kmeans++ algorithms consecutively. As our centroid is pre-defined, the number of iterations is constant in every test case. On the contrary, kmeans++ method works with random iterations. For getting insight into the clusters, we have plotted the cluster map in Figure \ref{fig:4} showing similar types of countries in terms of health care quality. The 2D plot also shows the cluster distributions in a two-dimensional space. Before jumping to the demonstration, it is worth mentioning that the model have been tested on the Intel\textsuperscript{\textregistered} Core\textsuperscript{TM} i7-8750H processor. 

\begin{itemize}
 \item[$\bullet$] Figure \ref{fig:3}(a) provides the iteration comparison for existing kmeans++ and our proposed method. When our proposed method have been applied, we find a constant number of iterations in each test. But the existing kmeans++ method’s iteration is random. It might be varied for different test cases. 
 \item[$\bullet$] Figure \ref{fig:3}(b) provides the execution time comparison for existing kmeans++ and our proposed method. For each test case, our model have executed the k-means clustering algorithm with the shortest time.    
\end{itemize}
Based on the experimental results, we claim that our model outperforms in the case of real-world application and reduces the computational power for the k-mean clustering algorithm.

\vspace{-2mm}
\section{Concluding Remarks}
\label{discussion}
\vspace{-2mm}
In this paper, we have presented an efficient clustering method that selects the optimal initial centroids of K-means clustering algorithm. With the help of the proposed method, we have efficiently created the clusters of different countries according to similar health care quality during COVID-19. While experimenting with COVID-19 datasets, our model outperforms in terms of a reduced number of constant iterations, that consequently reduces the execution times as well. Although our proposed model outperforms for real-world scenarios, it might be a little bit diverged if the number of instances of the dataset are high or extremely high. In the future, we have a plan to work on a huge number of instances and build an effective recommendation system.


\bibliographystyle{splncs04}
\bibliography{ICO}
\end{document}